\pdfoutput=1

\documentclass[11pt]{article}

\usepackage{EACL2023}

\usepackage{times}
\usepackage{latexsym}

\usepackage[T1]{fontenc}

\usepackage[utf8]{inputenc}

\usepackage{microtype}

\usepackage{amsmath}
\usepackage{ amssymb }
\usepackage[capitalise]{cleveref}

\usepackage{multicol}
\usepackage{multirow}

\DeclareMathOperator*{\argmax}{arg\,max}

\usepackage{graphicx}

\usepackage{xcolor}

\usepackage{booktabs}

\newcommand{\cinf}[1]{\scalebox{0.7}{\ensuremath{\pm #1}}}

\usepackage{ dsfont }

\newcommand{\expec}[1]{\ensuremath{\mathds{E}_{#1}}}

\newcommand{\marginal}[1]{\ensuremath{\Tilde{#1}}}

\usepackage{xspace}

\newcommand{\grammar}{}
\newcommand{\nogrammar}{\ensuremath{\ddagger}\xspace}

\newcommand{\frshort}{\textsc{f}$\rightarrow$\textsc{r}\xspace}
\newcommand{\rfshort}{\textsc{r}$\rightarrow$\textsc{f}\xspace}

\newcommand{\ar}{\textsc{ar}\xspace}
\usepackage{afterpage}

\newcommand{\ie}{i.e.\ }
\newcommand{\eg}{e.g.\ }

\usepackage{enumitem}

\setlist[itemize]{parsep=0pt,topsep=5pt,itemsep=0ex,partopsep=1ex,parsep=1ex}

\newcommand{\tightmath}{}

\usepackage[noend]{algpseudocode}
\usepackage{algorithm}

\title{Compositional Generalisation \\ with Structured Reordering and Fertility Layers}

\author{Matthias Lindemann$^1$ \and Alexander Koller$^2$ \and Ivan Titov$^{1,3}$ \\
$^1$ ILCC, University of Edinburgh,
$^2$ LST, Saarland University,
$^3$ ILLC, University of Amsterdam \\
{\small \texttt{m.m.lindemann@sms.ed.ac.uk}, \texttt{koller@coli.uni-saarland.de}, \texttt{ititov@inf.ed.ac.uk} }
}

\begin{document}
\maketitle

\begin{abstract}
Seq2seq models have been shown to struggle with compositional generalisation, \ie generalising to new and potentially more complex structures than seen during training.
Taking inspiration from grammar-based models that excel at compositional generalisation, we present a flexible end-to-end differentiable neural model that composes two structural operations: a fertility step, which we introduce in this work, and a reordering step based on previous work \citep{wang2021structured}. 
To ensure differentiability, we use the expected value of each step, which we compute using dynamic programming.
Our model outperforms seq2seq models by a wide margin on challenging compositional splits of realistic semantic parsing tasks that require generalisation to longer examples. It also compares favourably to other models targeting compositional generalisation.\footnote{\url{https://github.com/namednil/f-then-r}}

\end{abstract}

\section{Introduction}

Many NLP tasks require translating an input object such as a sentence into a structured output object such as a semantic parse. Recently, these tasks have been approached with seq2seq models with great success. 
However, seq2seq models also have been shown to struggle \textit{out-of-distribution} on compositional generalisation \citep{lake2018generalization, finegan-dollak-etal-2018-improving, kim-linzen-2020-cogs, hupkes2020compositionality}, \ie the model fails on examples that contain unseen compositions or deeper recursion of phenomena that it handles correctly in isolation.

\begin{figure}[t]
    \centering
    \includegraphics[width=\linewidth]{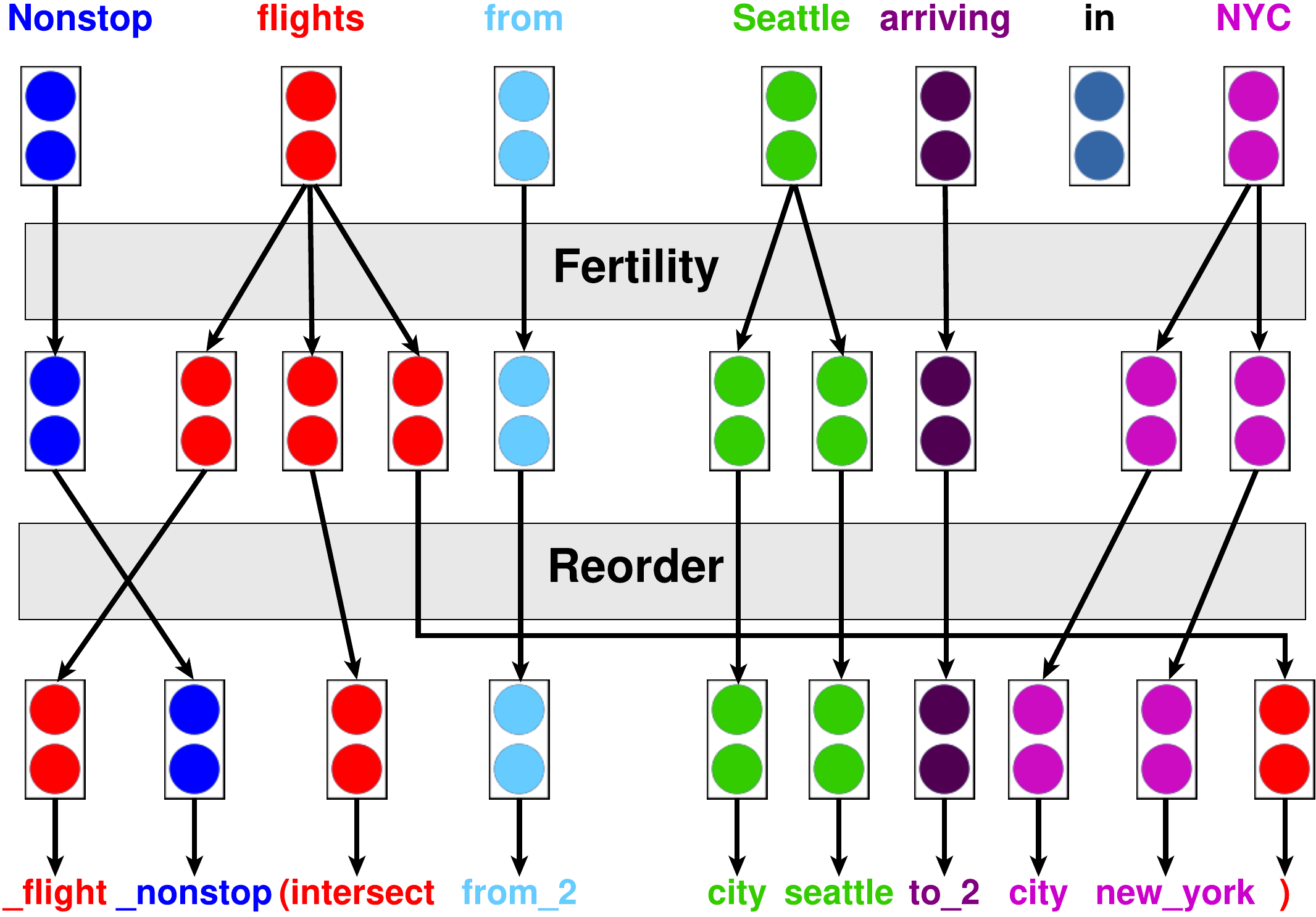}
    \caption{We model structural seq2seq tasks as the composition of differentiable fertility and phrase reordering layers. The model is trained end-to-end without direct supervision of the two structural layers.}
    \label{fig:first-example}
    \vspace{-5pt}
\end{figure}

Consider the example in \cref{fig:first-example}. Arguably, any model that produces the given semantic parse from the input in a generalisable way has to capture the correspondence between fragments of the input and fragments of the output, at least implicitly. This is challenging because of structural mismatches between input and output. For example, the fragment contributed by ``flights" is discontinuous and intertwined with the rest of the semantic parse.

In contrast to seq2seq models, grammar-based models such as synchronous context-free grammars (SCFGs) \citep{lewis1968syntax, chiang-2007-hierarchical} explicitly capture the compositional process behind the data and therefore perform very well in compositional generalisation setups. They can also model common structural mismatches like the one shown in the example. However, %
grammar-based models are rigid and brittle and thus do not scale well. %

In this work, we take inspiration from grammar-based models
and present an end-to-end differentiable neural model that is both flexible and generalises well compositionally. 
Our model consists of two structural layers: a phrase reordering layer originally introduced by \citet{wang2021structured} and a fertility layer, new in this work, which creates zero or more \textit{copies} of the representation of any input token. We show how to compose these layers to achieve strong generalisation.

We use a simple decoder and essentially translate each token after the fertility and reordering layers independently into one output token. We found this to lead to better generalisation than using an LSTM-based decoder. The simple decoder also makes it easy to integrate grammar constraints to ensure well-formed outputs, \eg in semantic parsing.

Most seq2seq models tend to predict the end of the sequence too early on instances that are longer than seen in training \citep{newman-etal-2020-eos}. Our model overcomes this by explicitly predicting the length of the output sequence as a sum of fertilities.

We first demonstrate the expressive capacity and inductive bias of our model on synthetic data. We then evaluate on three English semantic parsing datasets (Geoquery, ATIS, Okapi) and a style transfer task \citep{lyu-etal-2021-styleptb}. 
We clearly outperform seq2seq models both with and without pretraining %
in structural generalisation setups, particularly when the model has to generalise to longer examples.
Our model also compares favourably with existing approaches that target structural generalisation, and we obtain state-of-the-art results on the structural style transfer tasks.

To summarise, our main contributions are:
\begin{itemize}
    \item an efficient differentiable fertility layer;
    \item a flexible end-to-end differentiable model that composes two structural operations (fertility and reordering) and achieves strong performance in structural generalisation tasks.
\end{itemize}

\section{Related Work}

\paragraph{Fertility} The concept of fertility was introduced by \citet{brown-etal-1990-statistical} for statistical machine translation to capture that a word in one language is often consistently translated to a certain number of words in another language.
\citet{tu-etal-2016-modeling} and \citet{malaviya-etal-2018-sparse} incorporate fertility into the attention mechanism of seq2seq models.
\citet{cohn-etal-2016-incorporating, gu-etal-2016-incorporating} use heuristic supervision for training their fertility models.
In contrast to prior work, we learn an explicit fertility component jointly with the rest of our model.

\paragraph{Monotonic alignment} Related to fertility is the concept of monotonic alignment, \ie an alignment $a$ that maps output positions to input positions such that for any two output positions $i < j$, $a(i) \leq a(j)$. Monotonic alignments are usually modelled by an HMM-like model that places the monotonicity constraint on the transition matrix \citep{yu-etal-2016-online, wu-cotterell-2019-exact}, leading to a runtime of $O(|\textbf{x}|^2 |\textbf{y}|)$ with $\textbf{x}$ being the input and $\textbf{y}$ the output. \citet{raffel2017online} parameterise the alignment using a series of Bernoulli trials and obtain a training runtime of $O(|\textbf{x}| |\textbf{y}|)$. Our approach also has $O(|\textbf{x}| |\textbf{y}|)$ runtime.

\paragraph{Compositional generalisation} There is a growing body of work on improving the ability of neural models to generalise compositionally in semantic parsing. Good progress has been made in terms of generalisation to new lexical items  \citep{andreas-2020-good,akyurek-andreas-2021-lexicon, conklin-etal-2021-meta, csordas-etal-2021-devil, ontanon-etal-2022-making} but structural generalisation remains very challenging \citep{oren-etal-2020-improving, bogin-2022-local-structure}.

\citet{herzig-berant-2021-span} use a neural chart parser and induce latent trees with an approach similar to hard EM. Their model assumes that one input token corresponds to a single leaf in the tree. \citet{zheng-lapata-2022-disentangled} re-encode the input and partially generated output with a transformer for every decoding step to reduce entanglement and show considerable gains in structural generalisation.

There has also been work inspired by quasi-synchronous grammars (QCFGs, \citet{smith-eisner-2006-quasi}). \citet{shaw-etal-2021-compositional} heuristically induce a QCFG and create an ensemble of a QCFG-based parser and a seq2seq model. \citet{qiu-data-aug} use similar QCFGs for data augmentation for a seq2seq model. Our approach does not require constructing a grammar. Finally, \citet{kim-2021-nqscfg} introduces neural QCFGs which perform well on compositional generalisation tasks but are very compute-intensive.

Closest to our work is that of \citet{wang2021structured} who reorder phrases and use a monotonic attention mechanism on top. Our approach differs from theirs in several important aspects: (i) we use fertility instead of monotonic attention, which parameterises alignments differently; (ii) we apply the fertility step first and \textit{then} reorder the phrases, so our model can directly create alignments where output tokens aligned to the same input token do not have to be consecutive; (iii) we predict the length as the sum of the fertilities and not with an end-of-sequence token; (iv) they use an LSTM-based decoder network whereas we found that a simpler decoder can generalise better.

\section{Background}

\subsection{Structured Attention}
\label{seq:structured-att}

 We often want to model the relationship between input $\textbf{x}$ of length $n$ and output $\textbf{y}$ of length $l$ by means of a latent variable $Z$.
 In particular, we assume that $Z \in \mathcal{Z} \subseteq \mathds{B}^{n \times l}$ is a boolean alignment matrix. We want to predict $\textbf{y}$ from a representation produced by a function $g(Z, \textbf{x})$. In this case, the marginal distribution
 \begin{align}
     P(\textbf{y}|\textbf{x}) = \expec{P(Z|\textbf{x})} P(\textbf{y}|g(Z,\textbf{x})) \label{eq:intractable}
 \end{align}
 can be intractable to compute because $\mathcal{Z}$ often has exponential size in $\textbf{x}$. In some important cases however we can formulate a similar but tractable model by using \textit{structured attention} \citep{structured-att} which `pushes' the expectation inside the model:
 \begin{align*}
 P(\textbf{y}|\textbf{x}) \approx P(\textbf{y}|g(\marginal{Z},\textbf{x}))
 \end{align*}
 where $\marginal{Z} = \expec{P(Z|\textbf{x})} Z$ is now a `soft' rather than a boolean matrix. Note that $\marginal{Z}_{ij} = P(Z_{ij}=1|\textbf{x})$ is a marginal probability that often can be efficiently computed with a dynamic programme if $P(Z|\textbf{x})$ is factorisable. We will use such an approach for our model.

\subsection{Marginal Permutations}
\label{sec:bailin}
In this section, we briefly review the method of \citet{wang2021structured} that we use in our model.

\citet{wang2021structured} build on bracketing transduction grammars \citep{wu-1997-stochastic} and show how to compute a distribution over separable permutations \citep{BOSE1998277}. %
A permutation is separable, iff it can be represented as a permutation tree. Some permutations are not separable.
The internal nodes of a permutation tree are labelled as $\wedge$ or $\triangle$ and are interpreted as operations: $\wedge$ concatenates the values it receives from its left child with the value from its right child, whereas $\triangle$ concatenates them in reverse order. 
For example, the permutation tree $t = (\triangle \: (\wedge \: a \ b) \: (\triangle \: c \ d))$ represents the permutation $abcd \rightarrow dcab$.
Let $R_t(i,j)=1$ if the permutation described by tree $t$ maps position $i$ to position $j$, otherwise $R_t(i,j)=0$. %

\citet{wang2021structured} show how to compute the expected permutation matrix $\marginal{R}_{i,j} \triangleq \expec{P(t|\textbf{x})} R_t(i, j)$ in polynomial time with a CYK-style algorithm if $P(t|\textbf{x})$ factors according to the CYK chart. Crucially, the expected permutation can also be interpreted as a distribution over alignments, where $\marginal{R}_{i,j}$ is the \textit{marginal probability} that position $i$ aligns to position $j$. We will use this alignment distribution as a building block in our model.

\section{Overview of the Approach}
\label{sec:overview}
In this section, we give a general overview of our approach and defer the details to \cref{sec:fertility,sec:f-and-r}.

Conceptually, we want to model the transduction from input $\textbf{x}$ of length $n$ into output $\textbf{y}$ of length $l$ as the composition of two edit operations. First, we apply a \textit{fertility} step, in which we decide for each token what its fertility is, \ie how many \textit{copies}  we make of it. Assigning fertility of $0$ corresponds to deleting the token. 
This step yields an \textit{intermediate} sequence of tokens. We then reorder them using permutation trees (see \cref{sec:bailin}). In the last step, we individually translate these tokens into the output tokens. \cref{fig:first-example} shows an example where the fertility step and reordering apply to vector representations of tokens. 

The fertility step and the reordering can be represented as boolean matrices $F \in \mathds{B}^{n \times l}$ and $R \in \mathds{B}^{l \times l}$, respectively. These matrices denote alignments between the sequences before and after the operation. For example, $F_{i,j}=1$ means that input token $i$ aligns to intermediate token $j$, \ie $j$ is one of the copies of $i$. 

With this conceptualisation, we would ideally  use the following probabilistic model $P(\textbf{y}|\textbf{x})$:
\begin{flalign*}
P(\textbf{y}|\textbf{x}) = \expec{\underbrace{P(F|\textbf{x})}_{\text{Fertility}}} \big[  \expec{\underbrace{P(R|\textbf{x},F)}_{\text{Reordering}}} \underbrace{P(\textbf{y}|\textbf{x}, F, R}_{\text{Decoder}}) \big]
\end{flalign*}
At training time, the true fertility values are unknown but we observe the length $l$ of $\textbf{y}$, so we condition on it:
\begin{align}
P(\textbf{y}|\textbf{x}) = P(l|\textbf{x}) \cdot \expec{P(F, R| \textbf{x}, l)} P(\mathbf{y}|\textbf{x}, F, R) 
\end{align}
where $P(l|\textbf{x})$ can be computed with dynamic programming relying on the fertility model, as we explain in the next section.

Computing the marginal likelihood and the gradients is intractable.
 Instead of computing the likelihood exactly, one could sample and use a score function estimator \citep{williams1992simple} but the resulting gradient estimates have high variance.

Instead, we use structured attention as discussed in \cref{seq:structured-att} and `push' the expectations inside the model:
\begin{align}
 \expec{P(F, R| \textbf{x}, l)} P(\mathbf{y}|\mathbf{x}, F, R)   \approx P(\textbf{y}|\textbf{x}, \marginal{F}, \marginal{R})
\end{align}
with $\marginal{F} = \expec{P(F|\textbf{x},l)} F$ and $\marginal{R} = \expec{P(R|\textbf{x}, \marginal{F})}R$. 
$\marginal{F}$ and $\marginal{R}$  now represent `soft', differentiable versions of fertility and reordering. They approach their discrete counterparts as $P(F|\textbf{x},l)$ and $P(R|\marginal{F},\textbf{x})$ become peakier, which tends to happen over the course of training. 
We can view $(\marginal{F}\marginal{R})_{i,j} = \sum_{k} \marginal{F}_{i,k} \marginal{R}_{k,j}$ as a probability of aligning $i$ to $j$ in the composition of the operations.

A tendency to memorise larger chunks of the output might contribute to poor compositional generalisation \citep{hupkes2020compositionality}. In order to avoid this, we use a simple decoder that generates each output token independently. A first attempt might look like this:
\tightmath
\begin{align}
        P(\textbf{y}|\textbf{x},\marginal{F},\marginal{R}) = \prod_{i=1}^l \sum_{j, k} P(\textbf{y}_i|\textbf{x}_j) \marginal{F}_{j,k} \marginal{R}_{k,i} \label{eq:simplistic}
\end{align}
where the summation over $j$ and $k$ marginalises over all possible alignments to output position $i$.

\paragraph{Distinguishing copies}
The independence assumptions in \cref{eq:simplistic} imply that $P(\textbf{y}_i|\textbf{x}_j)$ will have the same distribution for \textit{all} copies of $x_j$, i.e. the model cannot express a preference to translate the first copy of \textit{Seattle} to \texttt{city} and the second copy to \texttt{seattle} in \cref{fig:first-example}. 
To enable this, we distinguish different copies of an input token.

We do this by defining $F$ not as a matrix but as a tensor $F \in \mathds{B}^{n \times l \times d}$ where $d$ is some fixed maximum fertility value. Let $F_{i,j,u}=1$ iff intermediate token $j$ is the $u$-th copy of input token $i$. For example, in \cref{fig:first-example}, $F_{4,6,1}=1$ and $F_{4,7,2}=1$.
With this definition of $F$ (and accordingly defined $\marginal{F}$), we can define a stronger decoder:
\begin{align*}
    P(\textbf{y}|\textbf{x},\marginal{F},\marginal{R}) = \prod_{i=1}^l \sum_{j,k,u} P(\textbf{y}_i|\textbf{x}_j, u) \marginal{F}_{j,k,u} \marginal{R}_{k,i} 
\end{align*}
where we now additionally marginalise over which copy of the input sequence we are translating. %

\section{Fertility and Alignment}
\label{sec:fertility}

In this section we describe how to compute $\marginal{F} = \expec{P(F|\textbf{x},l)} F$, i.e. the expected alignment that results from the fertility step given that the intermediate token sequence will have length $l$.

In the previous section, we looked at the fertility step mostly from the perspective of an alignment between the input tokens and the intermediate tokens. However, the fertility step is \textit{parameterised} as assigning a fertility value $f_i \in 0, \ldots, d \in \mathds{N}$ to every token $x_i$.  %
We now show how to compute $\marginal{F}$ efficiently as a function of the distribution over fertility values $P(\mathbf{f}|\textbf{x})$.

We denote the alignment that follows from $\mathbf{f}$ as $F(\mathbf{f})$, i.e. $F(\mathbf{f})_{i,j,u}=1$ iff intermediate token $j$ is the $u$-th copy of token $i$. $\marginal{F}$ can be expressed as:
\begin{align}
    \marginal{F} &= \expec{P(F(\mathbf{f})|\textbf{x}, l)} F(\mathbf{f}) \label{eq:def-fert-step}
\end{align}
where we assume that the fertility values are independent of each other conditioned on $\textbf{x}$:
\begin{align*}
   P(F(\mathbf{f})|\textbf{x}) \triangleq P(\mathbf{f}|\textbf{x}) \triangleq \prod_{i=1}^n P(f_i|\textbf{x})
\end{align*}
Note that conditioning on the output length $l$ in \cref{eq:def-fert-step} introduces inter-dependencies between the values. %
In order to compute $\marginal{F}_{i,j,u}$, we need to marginalise over all possible assignments to the fertility vector $\mathbf{f}$ which satisfy $F(\mathbf{f})_{i,j,u}=1$. We do this with a dynamic programming algorithm that is similar to the forward/backward algorithm for HMMs \citep{baum1972inequality}. %

\begin{figure}
    \centering
    \includegraphics[width=\linewidth]{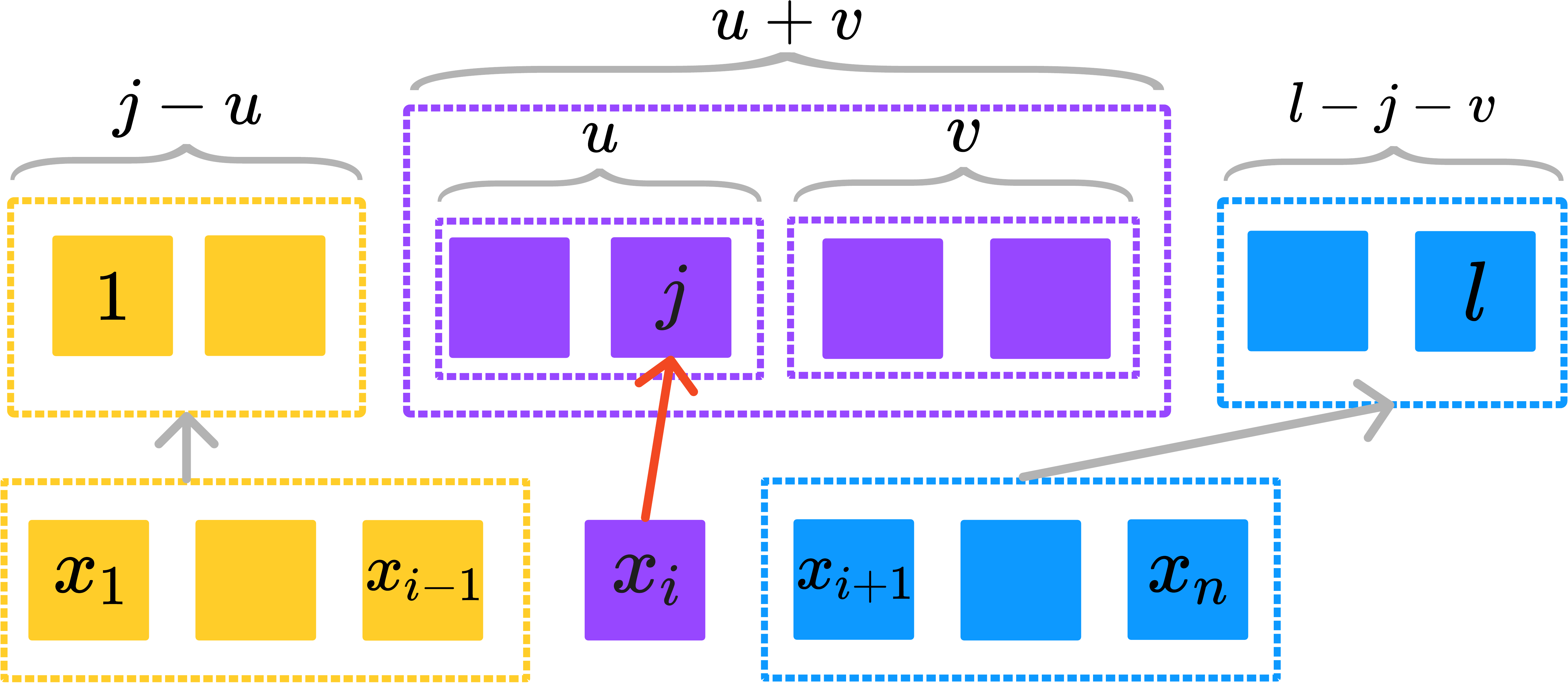}
    \caption{Efficiently computing the marginal probability that the $u$-th copy of $i$ is at position $j$. We partition the intermediate sequence into four parts, and marginalise over all possible ways of choosing $v$.}
    \label{fig:fertility}
    \vspace{-12pt}
\end{figure}

\paragraph{Computing marginals}
We characterise the situations where $F(\mathbf{f})_{i,j,u}=1$ as the integer solutions to a set of equations (see also \cref{fig:fertility}). First, in order for intermediate token $j$ to be the $u$-th copy of $i$, there should be $j-u$ intermediate tokens generated by input tokens preceding $i$:
\tightmath
\begin{flalign}
     f_1 +  \ldots + f_{i-1} &= j - u  \label{eq:before-i}
\end{flalign}
Second, the fertility $f_i$ has to be at least $u$. We capture this by requiring
\begin{align}
    f_i &= u + v  \label{eq:i}
\end{align}
with $v \geq 0$. Finally, the input tokens following $i$ have to contribute the remaining $l - j -v$ intermediate tokens:
\begin{align}
    f_{i+1} + \ldots + f_n &= l - j - v \label{eq:after-i}
\end{align}

We then compute the probability of creating a sequence of length $l$, where the $j$-th intermediate token is the $u$-th copy of $i$, by marginalising over $v$  (dropping $\textbf{x}$ for readability):
\begin{align}
    &P(F(\mathbf{f})_{i,j,u} = 1, f_0 + \ldots + f_{n+1} = l) \label{eq:mono-align} \\ 
    & = \sum_{v=0}^{\min(l-j,d-u)} P(f_0 + \ldots + f_{i-1}= j - u) \times \nonumber \\ & P(f_i = u + v) P(f_{i+1} + \ldots +f_{n+1} = l -j-v) \nonumber
\end{align}

We handle the boundary cases with $i=1, i=n$ by adding dummy variables $f_0$ and $f_{n+1}$ and setting $P(f_0 = 0) = P(f_{n+1} = 0) = 1$.

In order to compute \cref{eq:mono-align}, we also compute `forward' probabilities $P(f_0 + \ldots + f_i = h)$  and `backward' probabilities $P(f_i + \ldots + f_{n+1} = h)$ for all $i,h$. %
This can be done recursively:
\tightmath
\begin{align*}
& P(f_0 + \ldots + f_i = h) = \\
& \sum_{r=0}^d P(f_i = r) P(f_0 + \ldots + f_{i-1} = h-r)
\end{align*}
Finally, $ \marginal{F}_{i,j,u} = 1/ P(f_0 + \ldots + f_{n+1} = l) \times P(F(\mathbf{f})_{i,j,u}=1, f_0 + \ldots + f_{n+1} = l) $ because $\marginal{F}_{i,j,u}$ was defined in \cref{eq:def-fert-step} by conditioning on the length $l$. 
We provide pseudo-code for all steps in \cref{app:pseudo-code}.

\paragraph{Runtime} The runtime of the fertility step is dominated by computing \cref{eq:mono-align}. Note that $v$ has a range of at most $d$, thus, we can compute the probabilities for all $i,j,u$ in $O(n \cdot l \cdot d^2)$ time. 
\cref{eq:mono-align} can be computed in parallel for each $i,j,u$.

\section{Composing Fertility and Reordering}
\label{sec:f-and-r}
We will now describe in detail how $P(F|\textbf{x})$ and  $P(R|\textbf{x}, \marginal{F})$ are defined and how the fertility and reordering layers are composed.

\subsection{Fertility Layer}
The fertility layer takes a desired output length $l$ and a sequence of vectors $\mathbf{x}_1, \ldots, \mathbf{x}_n$ as input (GloVe embeddings \citep{pennington-etal-2014-glove} in our case) and returns a marginal alignment $\marginal{F}$ (see \cref{sec:fertility}) and a sequence of vectors $\mathbf{h}_1, \ldots, \mathbf{h}_l$ for use as input to the next layer. We compute the distribution over fertilities by first encoding $\mathbf{x}_1, \ldots, \mathbf{x_n}$ with a bidirectional LSTM, yielding a sequence of hidden states $\mathbf{h}^f_1, \ldots, \mathbf{h}^f_n$. We then model $P(f_i| \mathbf{x}) = \text{softmax}_{\tau}(\text{MLP}^f(\mathbf{h}^f_i))$, where $\tau$ is the temperature parameter of the softmax.

We use $\marginal{F}$ (\cref{eq:def-fert-step}) as a form of structured attention to compute the input to the reordering layer:
\begin{align*}
    \mathbf{h}_j =  \sum_{i,u} \marginal{F}_{i,j,u} (\mathbf{x}_i + \textbf{w}_u)
\end{align*}
where $\textbf{w}_u$ is a learned embedding indicating that $j$ is a $u$-th copy of some token. Intuitively, $\mathbf{h}_j$ for an intermediate token $j$ represents
the corresponding token in the input sequence and also indicates \textit{which} copy of that token it is.

\subsection{Reordering Layer}
Given the output $\mathbf{h}_1, \ldots, \mathbf{h}_l$ from the fertility layer, the reordering layer computes the alignment distribution $\marginal{R}$ as the expected permutation following \citet{wang2021structured}. This procedure involves populating a CYK-style chart with scores. We first run a bidirectional LSTM with a skip connection over $\mathbf{h}$ and compute a contextualised representation of the tokens after the fertility step:
\begin{align*}
    \mathbf{h}^r_i =  [ \text{LSTM}^r(\textbf{h}_{\leq i}), \text{LSTM}^r(\textbf{h}_{\geq i}) ] + \mathbf{h}_i
\end{align*}
Based on these representations, we compute scores for the chart following \citet{stern-etal-2017-minimal}.

\subsection{Decoder}
\label{sec:decoder}
Our decoder factors as follows (see \cref{sec:overview}):
\tightmath
\begin{align*}
    P(\textbf{y}|\textbf{x},\marginal{R},\marginal{F}) = \prod_{i=1}^l \underbrace{\sum_{j,k,u} P(\textbf{y}_i|\textbf{x}_j, u) \marginal{F}_{j,k,u} \marginal{R}_{k,i}}_{P(\textbf{y}_i|\textbf{x}, \marginal{R},\marginal{F})}
\end{align*}
$P(\textbf{y}_i|\textbf{x}_j, u)$ conditions only on the original input token and on the index indicating which copy of this token we are translating. For this reason, we contextualise the \textit{input} with a bidirectional LSTM with a skip connection: 
\begin{align}
    \textbf{h}'_j = \rho [\text{LSTM}^d(\mathbf{x}_{\leq j}), \text{LSTM}^d(\mathbf{x}_{\geq j})] + \mathbf{x}_j \label{eq:lstm-in-decoder}
\end{align}
with $\rho$ as hyperparameter.

We experiment with three versions of the decoder. In (i), we parameterise $P(\textbf{y}_i|\textbf{x}_j, u)$ as $ P(\textbf{y}_i|\textbf{x}_j, u) = \text{softmax}(W_u \text{MLP}(\textbf{h}'_j))$. In (ii), we additionally use a copy mechanism \citep{gu-etal-2016-incorporating}.
In (iii), we use an autoregressive variant where we encode $\textbf{y}_{<i}$ with an LSTM, defining $P(\textbf{y}_i|\textbf{y}_{<i}, u) = \text{softmax}(W_u \text{MLP}(\mathbf{h}'_j + \text{LSTM}(\textbf{y}_{<i}))$.

\subsection{Training}
As mentioned in \cref{sec:overview}, at training time we condition on the observed length $l$ of $\textbf{y}$:  $P(\textbf{y}|\textbf{x}) = P(l|\textbf{x}) P(\textbf{y}|\textbf{x}, \marginal{F}, \marginal{R}) $ with $\marginal{F}$ conditioned on $l$.
We use a weighted version of the log likelihood as the objective function:
\begin{align*}
\sum_i\lambda_1 \log P(l^i|\textbf{x}^i) + \log P(\textbf{y}^i|\textbf{x}^i,l^i)
\end{align*}
with $i$ ranging over the training examples.

For the semantic parsing tasks, we found it necessary to give our model a reasonable starting point in terms of alignments. We encourage it to respect high-confidence automatic alignments during the first $m$ training epochs by adding the following term to our objective function:
\begin{align*}
\lambda_2 \sum_{(i,j) \in \mathcal{A}} \log \sum_{k,u} \marginal{F}_{i,k,u} \marginal{R}_{k, j}
\end{align*}
where $\mathcal{A}$ is the set of alignments with a posterior probability of at least $\chi$ according to an IBM-1 alignment model \citep{brown-etal-1993-mathematics}.

Initialising an alignment model with alignments from a simpler model was a common strategy in statistical machine translation \citep{och2003systematic}. %

\subsection{Inference}
In order to make predictions with a trained model, we want to compute the most likely output $\textbf{y}$ given the input $\textbf{x}$. 
It is convenient to treat the length as a discrete variable 
and use the same algorithm for computing $\marginal{F}$ as derived for training. 
We therefore search for $\argmax_{\textbf{y}} P(\textbf{y}|\textbf{x}) = \argmax_{l} P(l|\textbf{x}) P(\textbf{y}^l|\textbf{x},l)$ with $\textbf{y}^l = \argmax_{\textbf{y}} P(\textbf{y}|\textbf{x}, \marginal{F}, \marginal{R})$. 
For any given $l$ we can easily find $\textbf{y}^l$ in versions (i) and (ii) of the decoder:
\begin{align*}
\textbf{y}^l_i = \argmax_{\textbf{y}_i} P(\textbf{y}_i|\textbf{x}, \marginal{R},\marginal{F}) 
\end{align*} %
For version (iii) of the decoder, we use greedy search instead. It would be too costly to compute $\textbf{y}^l$ for all $l$, so we explore only the top $k$ most likely lengths.

\paragraph{Grammar-based decoding} In executable semantic parsing, we want to produce only well-formed outputs, which can be characterised by a context-free grammar $G$. In practical applications, this grammar is needed to execute the query, so there is little extra engineering effort in using it for decoding. 
We search for $$\textbf{y}^l = \argmax_{\textbf{y} \in L(G)} P(\textbf{y}|\textbf{x}, l)$$
Because of the simple decoder we can do this exactly by applying a modified version of Viterbi CYK. Unlike in parsing though, the string is not observed because it is exactly what we are looking for. Therefore, we fill all entries from $i$ to $i$ in the chart $C$ with $C_{A, i,i} = \max_{A \rightarrow a \in G} P(y_i = a |\textbf{x}, \marginal{R},\marginal{F})$. We then continue with the normal Viterbi CYK with weights of 1 on all other rules.

\section{Evaluation}

\subsection{Synthetic Data}
\label{sec:mirroring}

In order to probe the expressive capacity and inductive bias of our model, we evaluate on mirroring task $T=\{(w, ww^R) | w \in \Sigma^* \}$, \eg $abc \rightarrow abccba$. 
The challenge is that the length of the dependency between output tokens grows with the length of the example.
Models are trained on examples with input lengths 3 to 9, and tested in two setups. In the first setup (Length), the model has to generalise to examples with lengths 11 to 20; we use examples with length 10 as validation data. In the second setup (unseen combination, UC), the model only sees the symbols \texttt{x}, \texttt{y} and \texttt{z} grouped together as \texttt{xyz} on the input side at training time. The model is tested on examples that contain \texttt{x,y} or \texttt{z} adjacent to other symbols. See \cref{seq:app-mirror} for further details on the setup. %

We compare our model without copying (\frshort) with a variant that first applies the reordering and then the fertility step (\rfshort) and autoregressive variants of the two (\ar \frshort and \ar \rfshort). As baselines, we also compare with an LSTM-based seq2seq model with attention and a Transformer with relative positional embeddings, which was previously shown by \citet{csordas-etal-2021-devil} to perform well at compositional generalisation. \ar \rfshort has similarities to \citet{wang2021structured} who first reorder and then use an autoregressive decoder with monotonic attention.

\begin{table}[t]
    \centering
\resizebox{\linewidth}{!}{
\begin{tabular}{lrrrrr|r}
\toprule
\textbf{Model} & \multicolumn{6}{c}{\textbf{Accuracy}} \\
 & \multicolumn{5}{c}{Length} & UC \\
& dev & 11 & 12 & 13 & 14 - 20 & test \\
\midrule
Transformer & 82.0 & 3.0 & 0.0 & 0.0 & 0.0 & 42.0 \\
LSTM & \textbf{100.0} & 94.3 & 67.9 & 6.9 & 0.0 & 0.1 \\
\midrule
\rfshort & 0.0 & 0.0 & 0.0 & 0.0  & 0.0 & 1.3 \\
\ar \rfshort & 90.0 & 85.7 & 89.8 & 87.5 & 80.5 & 1.2 \\
\midrule
\frshort & \textbf{100.0} & \textbf{100.0} & \textbf{100.0} & \textbf{100.0} & \textbf{100.0} & \textbf{79.9} \\
\ar \frshort &  \textbf{100.0} & \textbf{100.0} & \textbf{100.0} & \textbf{100.0} & \textbf{100.0} & 32.1 \\
\bottomrule
\end{tabular}}
    \caption{Exact match accuracy on the mirroring task.}
    \label{tab:mirror}
    \vspace{-8pt}
\end{table}

\paragraph{Results} \cref{tab:mirror} shows mean accuracy across 5 random initialisations. The accuracy of the relative Transformer and the LSTM-based seq2seq model drops sharply for longer inputs. In contrast, \frshort and \ar \frshort generalise perfectly even to much longer examples. In the UC setup, \frshort outperforms the rest by a wide margin. Interestingly, all autoregressive models struggle in this setup, including \ar \frshort which obtained perfect accuracy in the Length setup.
This is consistent with the hypothesis that autoregressive models tend to memorise entire chunks \citep{hupkes2020compositionality}.

\paragraph{Expressivity of \frshort and \rfshort} For this task, an input token corresponds to two output tokens that may be arbitrarily far apart from each other. \frshort can learn this alignment because this can be captured by a separable permutation (see \cref{sec:bailin}) following the fertility step duplicating every input token. In contrast, \rfshort and \ar\rfshort cannot represent the correct alignment directly because the fertility step is applied only after the reordering, leading to alignments between an input token and a \textit{contiguous span} in the output. However, in theory, they can represent this alignment \textit{implicitly} through the LSTM (\cref{eq:lstm-in-decoder}). %
The evaluation shows that this does not work reliably in practice: we find that \rfshort gets stuck in bad local minima and fails completely on the task. While \ar\rfshort performs well in the Length setup, it is weak in the UC setup.

\subsection{Geoquery}

Geoquery \citep{zelle1996learning} is a standard dataset for semantic parsing and has recently been used to evaluate to what extent semantic parsers are capable of generalising to (i) structurally unseen queries (template split), and (ii) structurally unseen long examples. We follow the setup of \citet{herzig-berant-2021-span}, using the variable-free FunQL representation \citep{Kate05learningto}, a copy mechanism, and evaluate with execution accuracy.

\begin{table}[t]
\resizebox{\linewidth}{!}{
\begin{tabular}{llll}
\toprule
\textbf{Model} & \textbf{iid} & \textbf{Template} & \textbf{Length} \\
\midrule
Seq2Seq \nogrammar (HB) & 78.5 & 46.0 & 24.3 \\
Seq2Derivation \grammar (HB) & 72.1 & 54.0 & 24.6 \\
BART-base\nogrammar (HB) & 87.1 & 67.0 & 19.3 \\
Span (HB) \grammar & 78.9 & 65.9 & 41.4 \\
Span + lexicon  (HB) \grammar & 86.1 & 82.2 & 63.6 \\
\citet{liu-etal-2021-learning-algebraic}\nogrammar & - & \textbf{84.1} & - \\
\citet{wang2021structured}\nogrammar & 75.2$^*$ & 43.2$^*$ & - \\
\midrule
\rfshort \grammar & \textbf{89.1}\cinf{1.0} & 80.4\cinf{1.2} & 68.6\cinf{1.4} \\
\rfshort \nogrammar & 83.5\cinf{0.7} & 73.0\cinf{1.6} & 49.2\cinf{5.1} \\
\midrule
\frshort \grammar & 88.6\cinf{3.3} & 79.9\cinf{2.5} & \textbf{68.8}\cinf{5.2} \\
\frshort \nogrammar & 80.7\cinf{2.3} & 68.7\cinf{4.3} & 53.4\cinf{5.9} \\
\ar \frshort \nogrammar & 81.1\cinf{1.0} & 52.8\cinf{3.3} & 37.6\cinf{1.8} \\
\bottomrule
\end{tabular}}
    \caption{Accuracy on different splits of Geoquery. HB is \citet{herzig-berant-2021-span} and \nogrammar refers to systems that do not enforce well-formedness of the output. $^*$ \citet{wang2021structured} use exact match accuracy and anonymise named entities instead of copying.}
    \label{tab:geo}
\end{table}

\paragraph{Results}
\cref{tab:geo} shows the results on the different splits. We report means and standard deviations of 5 random initialisations. 
Our method performs well across the different splits, and in particular on the length split that evaluates a challenging form of compositional generalisation. As an ablation, we remove the grammar-based decoding. We notice a considerable drop in accuracy %
but it still outperforms the baselines. The drop in accuracy is slightly bigger out-of-distribution than in-distribution.

In line with the experiments on synthetic data, \ar \frshort and the approach of \citet{wang2021structured} drastically lose accuracy when going from in-distribution to the compositional generalisation setups. This provides further evidence that a strong decoder can hinder compositional generalisation.

In contrast to the experiments on synthetic data, \rfshort and \frshort perform comparably. Manual inspection of the data shows that good alignments on this dataset can be obtained even with the stronger assumption on possible alignments made by \rfshort.

\subsection{ATIS}

ATIS \citep{dahl-etal-1994-expanding} is a semantic parsing datasets for flight bookings. In comparison to Geoquery, the queries tend to be longer and the word order is more flexible. We use the variable-free FunQL notation as annotated by \citet{guo-etal-2020-benchmarking-meaning}. Apart from the original iid test split, we create a length split: Semantic parses with fewer than 4 conjuncts form the training set, parses with exactly 4 conjuncts form the development set and the test set contains instances with more than 4 conjuncts. Details on the split and preprocessing are in \cref{app:data-atis}.

We compare our model with finetuned BART-base \citep{lewis2020bart}, an LSTM-based seq2seq model with attention and the relative Transformer \citep{csordas-etal-2021-devil}. We also run a version of the LSTM-based model with a large beam of size 50 and filter our instances that are not well-formed; the resulting outputs are well-formed at least $99.7\%$ of the time. 

\begin{table}[t]
\centering
\begin{tabular}{lll}
\toprule
\textbf{Model} & \textbf{iid} & \textbf{Length} \\
\midrule
LSTM seq2seq\grammar & 76.52\cinf{1.66} & 4.95\cinf{2.16} \\
LSTM seq2seq\nogrammar & 75.98\cinf{1.30} & 4.95\cinf{2.16} \\
Rel. Transformer\nogrammar & 75.76\cinf{1.43} & 1.15\cinf{1.41} \\
BART-base\nogrammar & \textbf{86.96}\cinf{1.26} & 19.03\cinf{4.57} \\
\midrule
\frshort \grammar & 74.15\cinf{1.35} & \textbf{35.41}\cinf{4.09} \\
\frshort \nogrammar & 68.26\cinf{1.53} & 29.91\cinf{2.91} \\
\bottomrule
\end{tabular}
\caption{Accuracy on different splits of ATIS.} %
    \label{tab:atis}
\end{table}

\paragraph{Results}
\cref{tab:atis} shows mean accuracy and standard deviations of 5 random initialisations. While on the iid split, our approach does not quite reach the same accuracy as the baselines, it outperforms them on the compositional length split by a margin of more than 16 points. Without grammar-based decoding, we again observe a noticeable loss in accuracy but we still substantially outperform BART on the length split. Constraining the output to be grammatical does not appear as beneficial for the LSTM baseline.

\subsection{Okapi}

\begin{table}[t]
    \centering
    \resizebox{\linewidth}{!}{
    \begin{tabular}{lrrr}
    \toprule
    \textbf{Model} & \textbf{Calendar} & \textbf{Document} & \textbf{Email} \\
    \midrule
BART-base\nogrammar & 36.7\cinf{3.0} & 2.7\cinf{2.1} & 20.5\cinf{9.8} \\
\frshort\grammar & \textbf{69.5}\cinf{13.9} & \textbf{42.4}\cinf{5.7} & \textbf{55.6}\cinf{2.7} \\
\frshort\nogrammar & 57.2\cinf{19.9} & 36.1\cinf{5.6} & 43.9\cinf{3.8} \\
    \bottomrule
    \end{tabular}
    }
    \caption{Accuracy on length splits by domain on Okapi.}
    \label{tab:okapi}
\end{table}

\begin{figure}[t]
    \includegraphics[width=\linewidth]{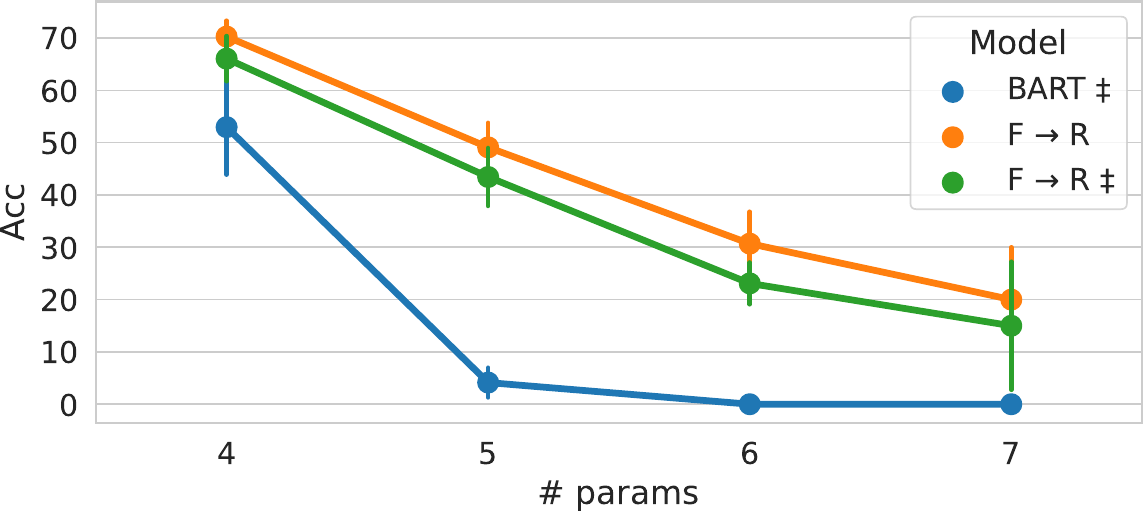}
    \captionof{figure}{Accuracy on the document domain of Okapi by number of parameters in the gold logical form.}
    \label{fig:okapi-doc-by-length}
\end{figure}

\citet{okapi} introduce a semantic parsing dataset for evaluating compositional generalisation on three domains (document, calendar, email). Since template splits were not found to be challenging, we focus on generalising to longer examples. Models are trained on short examples with up to 3 `parameters' (such as filtering based on an attribute or ordering results) and are tested on examples with more parameters (at least 4 for the calendar and email domain and at least 5 for the document domain). The splits are described in \cref{app:data-okapi}.  
In contrast to the other datasets we consider, Okapi is noisy because it was collected with crowd workers. This presents an additional challenge.

\paragraph{Results} \cref{tab:okapi} shows the results of our model with copying from 5 random initialisations. \frshort outperforms fine-tuned BART-base by a large margin, in particular on the challenging split of the document domain.

\cref{fig:okapi-doc-by-length} shows the accuracy on the development and test set of the document domain as a function of the number of parameters in the gold logical form. BART performs relatively well when applied to examples with one more parameter than seen in the training set but then its performance drops sharply. \frshort is more accurate and its accuracy also drops much slower with the number of parameters. We notice different failure modes for the two models: on the test set, BART deviates by 5.4 tokens on average from the gold length, whereas \frshort deviates only by 1.0 tokens on average. This is in line with the observations of \citet{newman-etal-2020-eos} that seq2seq models systematically predict the end of sequence token too early on long out-of-distribution examples. Our results suggest that predicting length as a sum of fertilities is more robust towards this shift in distribution.

\subsection{Style Transfer}

\begin{table*}[t]
\resizebox{\linewidth}{!}{
\begin{tabular}{lllllllll}
\toprule
\textbf{Transfer Type} & \textbf{Approach} & \textbf{BLEU-1} & \textbf{BLEU-2} & \textbf{BLEU-3} & \textbf{BLEU-4} & \textbf{METEOR} & \textbf{ROUGE-L} & \textbf{CIDEr} \\
\midrule
\multirow{5}{*}{ Active to passive } & GPT-2 \citep{lyu-etal-2021-styleptb} & 47.6 & 32.9 & 23.8 & 18.9 & 21.6 & 46.4 & 1.820 \\
& Seq2seq \citep{kim-2021-nqscfg} & 83.8 & 73.5 & 67.3 & 59.8 & 46.7 & 77.1 & 5.941 \\
& Neural QCFG \citep{kim-2021-nqscfg} & 83.6 & 77.1 & 71.3 & 66.2 & 49.9 & 80.3 & 6.410 \\
& \textcolor{gray}{Human} \citep{lyu-etal-2021-styleptb} & \textcolor{gray}{93.1} & \textcolor{gray}{88.1} & \textcolor{gray}{83.5} & \textcolor{gray}{79.5} & \textcolor{gray}{58.7} & \textcolor{gray}{90.5} & \textcolor{gray}{8.603} \\
& \frshort & \textbf{92.1}\cinf{1.3} & \textbf{85.4}\cinf{1.5} & \textbf{79.7}\cinf{1.7} & \textbf{74.6}\cinf{1.9} & \textbf{55.9}\cinf{0.5} & \textbf{86.0}\cinf{1.1} & \textbf{7.610}\cinf{0.130} \\
\midrule
\multirow{5}{*}{ Adj. emphasis } & GPT-2 \citep{lyu-etal-2021-styleptb} & 26.3 & 7.9 & 2.8 & 0.0 & 11.2 & 18.8 & 0.386 \\
& Seq2seq \citep{kim-2021-nqscfg} & 50.5 & 29.6 & 18.4 & 11.9 & 24.2 & 51.4 & 1.839 \\
& Neural QCFG \citep{kim-2021-nqscfg} & 67.6 & 50.6 & 39.3 & 31.6 & 37.3 & 68.3 & 3.424 \\
& \textcolor{gray}{Human} & \textcolor{gray}{83.4} & \textcolor{gray}{75.3} & \textcolor{gray}{67.9} & \textcolor{gray}{66.1} & \textcolor{gray}{52.2} & \textcolor{gray}{81.1} & \textcolor{gray}{6.796} \\
& \frshort  & \textbf{78.3}\cinf{1.4} & \textbf{61.9}\cinf{0.7} & \textbf{49.9}\cinf{1.3} & \textbf{40.5}\cinf{1.4} & \textbf{43.0}\cinf{1.0} & \textbf{69.1}\cinf{0.6} & \textbf{4.268}\cinf{0.170} \\
\midrule
\multirow{5}{*}{ Verb emphasis } & GPT-2 \citep{lyu-etal-2021-styleptb} & 30.9 & 17.0 & 9.5 & 4.1 & 14.0 & 29.2 & 0.593 \\
& Seq2seq \citep{kim-2021-nqscfg} & 52.6 & 38.9 & 29.4 & 21.4 & 29.4 & 46.4 & 2.346 \\
& Neural QCFG \citep{kim-2021-nqscfg} & 66.4 & 51.2 & 40.7 & 31.9 & 37.0 & \textbf{58.9} & 3.227 \\
& \textcolor{gray}{Human} \citep{lyu-etal-2021-styleptb} & \textcolor{gray}{64.9} & \textcolor{gray}{56.9} & \textcolor{gray}{49.3} & \textcolor{gray}{42.1} & \textcolor{gray}{43.3} & \textcolor{gray}{69.3} & \textcolor{gray}{5.668} \\
& \frshort & \textbf{68.4}\cinf{0.6} & \textbf{52.7}\cinf{0.6} & \textbf{41.6}\cinf{0.7} & \textbf{32.8}\cinf{0.6} & \textbf{37.4}\cinf{0.4} & \textbf{58.9}\cinf{0.4} & \textbf{3.498}\cinf{0.121} \\
\bottomrule
\end{tabular}
}
    \caption{Results on the hard style transfer tasks from \citet{lyu-etal-2021-styleptb}. All models except for GPT2 use copying.}
    \label{tab:style-ptb}
\end{table*}

In addition to being important for compositional generalisation, structural inductive biases can help when only little data is available. We evaluate our model in such a scenario on the style transfer tasks of \citet{lyu-etal-2021-styleptb}. A model is given an English sentence and asked to reformulate it to conform with a certain `style'. 
We focus on the tasks identified as challenging by \citet{lyu-etal-2021-styleptb}: \textit{active to passive} (2462 training examples), \textit{adjective emphasis} (627 examples) and \textit{verb emphasis} (1081 examples).

For the emphasis tasks, the word to be emphasised is provided in the input. Following \citet{kim-2021-nqscfg}, to incorporate it, we add a special learned embedding vector to the embedding of that token.

\paragraph{Results} \cref{tab:style-ptb} shows the results comparing our \frshort model to previous work based on three random initialisations. We achieve state-of-the-art results on all style transfer tasks on all metrics. The improvement compared to prior work is strongest for the active-to-passive task and weakest for the verb emphasis task, where our model ties with \citet{kim-2021-nqscfg} in terms of ROUGE-L.

\section{Conclusion}
We presented a flexible end-to-end differentiable model for structured NLP tasks. It predicts the output sequence from the input by composing a fertility layer with a reordering layer. The evaluation shows that our model performs well in structural generalisation setups, in particular when the model has to generalise to longer examples than seen during training. In contrast, the accuracy of standard seq2seq models drops sharply on longer examples.

The efficient fertility layer introduced in this work may be useful in other scenarios as well, \eg in non-autoregressive machine translation, or for (unsupervised) sentence compression when the fertility is restricted to 0 or 1. Future work could also investigate other structured layers and the best ways of composing and training them.

\section{Limitations}
The fertility layer is efficient but a limitation of the model presented in this paper is the high runtime complexity of the reordering layer. %
It makes it impractical for long output sequences (\eg more than 50 tokens). While the structured reordering step can represent many permutations of practical interest, we observed a small number of cases where our model could not produce the correct permutation. 
We note that our approach is modular and the reordering layer could be replaced by a faster one with fewer restrictions in future work, \eg based on one-to-one matchings.

While our method obtains strong accuracy in compositional generalisation setups without contextualised encoders, it remains an open question how different ways of integrating contextualised encoders affect the performance of our method in compositional generalisation setups. In addition, it is an open question how a \textit{pretrained decoder} would influence our model's ability to generalise compositionally since we found that a non-pre-trained LSTM-based decoder can be detrimental.

\section*{Acknowledgements}
We thank Bailin Wang and Jonas Groschwitz for technical discussions and we thank Agostina Calabrese and Verna Dankers for comments on this paper.

ML is supported by the UKRI Centre for Doctoral Training in Natural Language Processing, funded by the UKRI (grant EP/S022481/1), the University of Edinburgh, School of Informatics and School of Philosophy, Psychology \& Language Sciences, and a grant from Huawei Technologies. ML and IT acknowledge the support of the European Research Council (ERC StG BroadSem 678254). IT is also supported by the Dutch National Science Foundation (NWO Vici VI.C.212.053).

\bibliography{anthology,custom}
\bibliographystyle{acl_natbib}

\appendix

\newpage

\renewcommand{\floatpagefraction}{1.0}
\renewcommand{\dbltopfraction}{1.0}
\renewcommand{\topfraction}{1.0}
\renewcommand{\floatpagefraction}{1.0}

\section{Pseudo code}
\label{app:pseudo-code}

We present the algorithms for computing $\marginal{F}$ in python-style pseudo code, using 0-based indexing and \texttt{range(i,k)} refers to the integers $i, \ldots, k-1$. As in the main paper, $d$ refers to the maximum fertility value (hyperparameter).

\cref{alg:forward} shows how to compute all forward probabilities. If \texttt{rev(f)} reverses matrix \texttt{f} on the first dimension (e.g. like \texttt{torch.flip}), then we can compute the backward probabilities as \texttt{rev(fwd(rev(f), l))}. \cref{alg:marginals} shows how $\marginal{F}$ can be computed.

\algrenewcommand\algorithmicindent{1.0em}%

\begin{algorithm}[t]
	\caption{Forward}
	\begin{algorithmic}[1] 
	    \Require f $\in \mathbb{R}^{n \times d}$, normalised over the second axis, output sequence length $l$ 
		\Function{\texttt{fwd}}{f, $l$}
		\State Init fwd $\in \mathbb{R}^{n \times l}$ with zeros
		\For{$h$ in range(0, d)} \Comment Base case
    		 \State fwd[0, h] = f[0, h]
    	\EndFor
		\For{$i$ in range(1, n)} \Comment Recursion
		    \For{$h$ in range(0, $l$)}
    		    \For{$r$ in range(0, min(d, h+1))}
    		        \State fwd[i, h] += fwd[i-1, h-r] $\times$ f[i, r]
    		    \EndFor
    		 \EndFor
		\EndFor
		\State \textbf{return} fwd
		\EndFunction
	\end{algorithmic}
	\label{alg:forward}
\end{algorithm}

\begin{algorithm}[t]
	\caption{Marginals}
	\begin{algorithmic}[1] 
	    \Require f $\in \mathbb{R}^{n \times d}$, normalised over the second axis, output sequence length $l$
		\Function{\texttt{marginals}}{$f, l$}
		\State Init F $\in \mathbb{R}^{n \times l \times d}$ with zeros
		\State fwd = \texttt{fwd}(f, $l$)
		\State bwd = \texttt{rev(fwd(rev(f),$l$))}
		\For{$i$ in range(0, n-1)}
		    \For{$j$ in range(0, $l$)}
		        \For{$u$ in range(1, min(d, j+1+1))}
		            \For{$v$ in range(0, min(d-u, $l$-j))}
		                \State F[i, j, u] += fwd[i, j+1-r1] $\times$
		                \State $\quad$ f[i,u+v]$\times$bwd[i+1,$l$-(j+1+v)]
		            \EndFor
		        \EndFor
		    \EndFor
		 \EndFor
		\State divide all entries in F by fwd[i, $l$]
		\State \textbf{return} F
		\EndFunction
	\end{algorithmic}
	\label{alg:marginals}
\end{algorithm}

\section{Data, grammars, pre- and post-processing}

\subsection{Synthetic data}
\label{seq:app-mirror}
In both setups, we generate 4000 training examples, 200 development examples and 1000 test examples. We use an alphabet size of $|\Sigma| = 11$ for the Length setup and  $|\Sigma| = 11 + 3$ for the UC setup to accommodate for \texttt{x,y,z}. Symbols are chosen uniformly at random. In the Length setup, we choose the length of the example uniformly at random. In the UC setup, we do so as well but with probability $0.2$ we insert an \texttt{xyz} cluster if this does not exceed the maximum length.

In the UC setup, we use development data from the training distribution. 

We do not use grammar-based decoding or copying on the synthetic data.

\subsection{Geoquery}
We remove all parentheses in the logical forms, as they can be restored in post-processing.
We also remove quotes around named entities in preprocessing to enable copying (\texttt{'texas'} becomes \texttt{texas}) and restore them in post-processing. Following \citet{herzig-berant-2021-span}, we only allow copying of named entities and do not copy predicate symbols (e.g. \texttt{river}).

We use the grammar of \citet{wong-mooney-2006-learning} for decoding as provided by \citet{guo-etal-2020-benchmarking-meaning}.

\subsection{ATIS}
\label{app:data-atis}
We found that a naive length split led to having very few examples in the training set that used a date since both the month and the day count as one conjunct each. Therefore, we created 33 templates with three conjuncts based on existing ATIS examples (with four or more conjuncts) that contain dates and add 8 instances of each template with randomly chosen dates to the training set. In addition, we removed any exact duplicate samples from the data.

Similarly to Geoquery, we remove those parentheses that can be deterministically recovered in post-processing. However, in contrast to Geoquery the parentheses for \texttt{or} and \texttt{intersection} need to be kept because the arities of those operators are not fixed. We run all our experiments on this representation of ATIS.

For grammar-based decoding we use the ``typed" grammar provided by \citet{guo-etal-2020-benchmarking-meaning} and do not use the copy mechanism.

\subsection{Okapi}
\label{app:data-okapi}

We found there was too much distributional overlap in the original length split provided by \citet{okapi} and therefore use our own split:

For the document domain, our development set contains examples with 4 parameters, and the test set contains examples with at least 5 parameters. For the other two domains, there is insufficient data for such out-of-distribution development sets. Therefore, we chose our test set to contain all examples with at least 4 parameters and our development sets to consist of 95\% in-distribution data and 5\% of examples from the the examples with 4 parameters (which makes the bulk of the test distribution). %

We manually create a grammar of well-formed logical forms for the three domains of Okapi (included in the code).

\subsection{StylePTB}
For comparability, we also tokenise on whitespace following \citet{kim-2021-nqscfg}. We do not restrict the output of the model with a grammar.

\begin{table}[t]
    \centering
\resizebox{\linewidth}{!}{
\begin{tabular}{llrrr}
\toprule
\textbf{Dataset} & \textbf{Split/Version} & \textbf{Train} & \textbf{Dev} & \textbf{Test} \\
\midrule
\multirow{3}{*}{Geoquery } & iid & 540 & 60 & 280 \\
& template & 544 & 60 & 276 \\
& length & 540 & 60 & 280 \\
\midrule
\multirow{2}{*}{ATIS } & iid & 4465 & 497 & 448 \\
& length & 4017 & 942 & 331 \\
\midrule
\multirow{3}{*}{Okapi/length } & Calendar & 1145 & 200 & 1061 \\
& Document & 2328 & 412 & 514 \\
& Email & 2343 & 200 & 991 \\
\midrule
\multirow{3}{*}{Style transfer } & active to passive & 2462 & 136 & 137 \\
& adjective emphasis & 627 & 34 & 35 \\
& verb emphasis & 1081 & 60 & 60 \\
\bottomrule
\end{tabular}
}
    \caption{Number of examples per dataset/split.}
    \label{tab:num-examples}
\end{table}

\section{Details on evaluation metrics}
We provide code for all evaluation metrics in our repository/dependencies.

\paragraph{Geoquery}
We use the code of \citet{herzig-berant-2021-span} to compute execution accuracy (\url{https://github.com/jonathanherzig/span-based-sp}). 

\paragraph{ATIS}
We allow for different order of conjuncts between system output and gold parse in computing accuracy. We do this by sorting conjuncts before comparing two trees node by node.

\paragraph{Okapi}
We follow \citet{okapi} and disregard the order of the parameters for computing accuracy. Since \citet{okapi} did not make their evaluation code publicly available, we use our own implementation. Our implementation uses sets and does not punish a model for predicting a correct parameter multiple times. For example, if the gold logical form contains \texttt{FILTER message.isRead eq False}, a necessary condition for a prediction to count as correct is that it must contain this string \textit{at least} once.

\paragraph{StylePTB}
We follow \citet{kim-2021-nqscfg} and use \url{https://github.com/Maluuba/nlg-eval} (commit \texttt{7f79930}) for all evaluation metrics, ensuring that BLEU, ROUGE and METEOR are scaled to 0 - 100.

\section{Hyperparameters}

We provide a configuration file for each of our models with the chosen hyperparameters in our code repository (\texttt{configs/}). We set the maximum fertility value $d$ to $d=4$ for all datasets except for the style transfer tasks where we set it to $d=3$. 

At test time, we explore the top $k=5$ most likely lengths when using grammar-based decoding. Without grammar-based decoding we used $k=1$ as using $k=5$ provided little improvement.

Many but not all instances of Geoquery require identity permutations. We found this to lead to the issue that the model gets stuck in a very steep local minimum within the first epoch where it would predict only identity permutations. We fixed this issue by reducing the learning rate in the feedforward network that predicts the scores for the permutation trees to $1 \times 10^{-6}$.

\subsection{Hyperparameter selection}

We select hyperparameters using a combination of manual selection and a random search. We optimise hyperparameters for accuracy on the development set of compositional generalisation splits, where available, (Length setup for the synthetic data, template split for Geoquery), and then use those hyperparameters for all splits of a (domain of a) dataset.

The high variance in accuracy across random initialisations often observed in compositional generalisation setups makes it difficult to tune hyperparameters even if an out-of-distribution development set exists. We restrict random hyperparameter search to two random seeds. After the hyperparameter search, we pick the two most promising configurations (according to (execution) accuracy), pick a new random seed and train them again to choose the one which provides the most stable accuracy (approximated as the highest accuracy on the new seed). We then run the main experiments with the chosen hyperparameters and completely new random seeds.

We randomly sample 20 configurations per hyperparameter search. Since this procedure is expensive, we do not run train our models fully to convergence. The bounds of the hyperparameter search are reported in our repository. 

For all our models, we initially chose hyperparameters manually, and then ran a random hyperparameter search as described above. If the manually chosen hyperparameters resulted in same or better performance (on the development set) on average, we kept those, and otherwise used the ones found by the hyperparameter search. 

For example, on Geoquery, we noticed a particular sensitivity to hyperparameters, and the manually selected hyperparameters for \frshort performed best with low variance, whereas for \rfshort, the hyperparameters found by the random search were better than the manually chosen ones. We think this sensitivity is at least in part caused by the small size of the dataset.

\paragraph{Style transfer} In contrast to the other tasks we evaluate on, we did not run a hyperparameter search for the style transfer tasks and use the same manually determined hyperparameters for \frshort across all style transfer tasks.

\begin{table}[]
    \centering
    \resizebox{\linewidth}{!}{
\begin{tabular}{llr}
\toprule
\textbf{Dataset} & \textbf{Model} & \textbf{\# params} \\
\midrule
\multirow{4}{*}{Mirror } & \frshort and \rfshort & 1.019 \\
& \ar \frshort and \ar \rfshort & 1.12 \\
& LSTM & 3.187 \\
& Transformer & 33.067 \\
\midrule
\multirow{2}{*}{Geoquery } & \frshort and \rfshort & 2.541 \\
& \ar \frshort & 2.765 \\
\midrule
\multirow{3}{*}{ATIS } & LSTM & 4.669 \\
& Transformer & 58.468 \\
& \frshort & 3.511 \\
\midrule
Okapi/Calendar & \frshort & 2.466 \\
Okapi/Document & \frshort & 2.468 \\
Okapi/Email & \frshort & 2.485 \\
\midrule
Style/Active to passive & \frshort & 2.814 \\
Style/Adjective emphasis & \frshort & 2.698 \\
Style/Verb emphasis & \frshort & 3.133 \\
\bottomrule
\end{tabular}
}
    \caption{Number of parameters in millions in our models.}
    \label{tab:my_label}
\end{table}

\section{Computing infrastructure and runtime}

All experiments were run on GeForce GTX 1080 Ti or GeForce GTX 2080 Ti with 12GB RAM and Intel Xeon Silver or Xeon E5 CPUs.

The runtime of one run contains the time for training, evaluation on the devset after each epoch and running the model on the test set. We show the runtime the models we train in \cref{tab:runtimes}.

\begin{table}[t]
\resizebox{\linewidth}{!}{
\begin{tabular}{llll}
\toprule
\textbf{Dataset} &\textbf{Model} & \textbf{Epochs} & \textbf{Runtime} \\
\midrule
\multirow{6}{*}{Mirror } & \frshort & 7 & 8 min \\
& \rfshort & 20 & 8 min \\
& \ar\frshort & 20 & 30 min \\
& \ar\rfshort & 20 & 20 min \\
& Transformer & 200 & 15 min \\
& LSTM & 60 & 4 min \\
\midrule
Geo & \frshort & 100 & 20 min \\
\midrule
\multirow{4}{*}{ATIS } & \frshort & 100 & 11-12h \\
& Transformer & 20 & 10 min \\
& LSTM & 20 & 10 min \\
& BART & 50 & 1.3h \\
\midrule
\multirow{2}{*}{Okapi / Calendar } & \frshort & 70 & 1 h \\
& BART & 40 & 15 min \\
\multirow{2}{*}{Okapi / Document } & \frshort & 70 & 1.5 h \\
& BART & 60 & 30 min \\
\multirow{2}{*}{Okapi / Email } & \frshort & 70 & 1.5 h \\
& BART & 40 & 30 min \\
\midrule
Active $\rightarrow$ Passive & \frshort & 60 & 1 h \\
Adj. emphasis & \frshort & 60 & 20 min \\
Verb emphasis & \frshort & 60 & 30 min \\
\bottomrule
\end{tabular}
}
\caption{Average total runtime of the models we train. For comparison on Geoquery, \citet{herzig-berant-2021-span} report a runtime of 2 hours on comparable hardware to ours.}
\label{tab:runtimes}
\end{table}

\section{Additional results}
\label{seq:app-results-geo}

\paragraph{Geoquery}
\cref{tab:geo-exact} shows \textit{exact match} accuracy of our models for comparison and \cref{tab:geo-dev} shows results on the development set.

\paragraph{ATIS} \cref{tab:atis-dev} shows results on the development set. \cref{tab:length-atis} shows the average deviation from the gold length.

\paragraph{Okapi} \cref{tab:okapi-dev} shows results on the development set.

\paragraph{StylePTB} 
\cref{tab:style-ptb-dev} shows results on the development set.

\begin{table*}
\resizebox{\linewidth}{!}{
	\begin{tabular}{cccccccc}
		\toprule
		\textbf{Transfer Type} & \textbf{BLEU-1} & \textbf{BLEU-2} & \textbf{BLEU-3} & \textbf{BLEU-4} & \textbf{METEOR} & \textbf{ROUGE-L} & \textbf{CIDEr} \\
		\midrule
		Active to passive & 94.1\cinf{0.3} & 88.6\cinf{0.2} & 83.8\cinf{0.2} & 79.7\cinf{0.3} & 57.6\cinf{0.1} & 87.9\cinf{0.5} & 7.883\cinf{0.061} \\
		Adj. emphasis & 78.4\cinf{1.2} & 64.3\cinf{1.0} & 54.5\cinf{0.7} & 46.9\cinf{0.9} & 44.0\cinf{0.4} & 69.5\cinf{0.6} & 4.809\cinf{0.113} \\
		Verb emphasis & 67.6\cinf{0.5} & 51.2\cinf{0.7} & 40.6\cinf{0.9} & 32.7\cinf{1.3} & 36.8\cinf{0.3} & 59.1\cinf{0.7} & 3.587\cinf{0.118} \\
		\bottomrule
	\end{tabular}}
	\caption{Development accuracy of our \frshort model with copying on StylePTB. We report means and standard deviations of three random initialisations.}
	\label{tab:style-ptb-dev}
\end{table*}

\begin{table}[t]
\begin{tabular}{llll}
\toprule
\textbf{Model} & \textbf{iid} & \textbf{Template} & \textbf{Length} \\
\midrule
\rfshort\grammar & 83.5\cinf{0.7} & 72.7\cinf{2.0} & 54.0\cinf{4.0} \\
\rfshort\nogrammar & 76.4\cinf{1.4} & 65.8\cinf{2.2} & 39.4\cinf{6.2} \\
\frshort\grammar & 83.4\cinf{2.6} & 72.0\cinf{3.3} & 55.2\cinf{5.3} \\
\frshort\nogrammar & 76.9\cinf{2.2} & 63.6\cinf{4.7} & 46.0\cinf{6.6} \\
\ar \frshort\nogrammar & 77.4\cinf{1.3} & 47.5\cinf{4.2} & 34.4\cinf{2.2} \\
\bottomrule
\end{tabular}

\caption{Exact match accuracy on the splits of \textbf{Geoquery}. We report mean and standard deviation of the 5 random initialisations shown in the main paper.}
\label{tab:geo-exact}
\end{table}

\begin{table}[t]
	\begin{tabular}{lrrr}
		\toprule
		\textbf{Model} & \textbf{iid} & \textbf{Template} & \textbf{Length} \\
		\midrule
\frshort\nogrammar & 84.3\cinf{3.0} & 85.3\cinf{1.4} & 83.7\cinf{1.4} \\
\rfshort\nogrammar & 83.7\cinf{2.2} & 83.0\cinf{3.0} & 79.0\cinf{3.2} \\
\ar \frshort\nogrammar & 83.7\cinf{2.7} & 81.7\cinf{2.0} & 87.3\cinf{2.5} \\
\bottomrule
	\end{tabular}
	\caption{Mean and standard deviations of execution accuracy on the development sets of the \textbf{Geoquery} splits (without grammar-based decoding).}
	\label{tab:geo-dev}
\end{table}

\begin{table}[t]
    \begin{tabular}{lrr}
    \toprule
    \textbf{Model} & \textbf{iid} & \textbf{Length} \\
    \midrule
    LSTM seq2seq \nogrammar & 0.46\cinf{0.09} & 5.39\cinf{0.42} \\
    Rel. Transformer \nogrammar & 0.49\cinf{0.07} & 6.19\cinf{0.52} \\
    BART-base \nogrammar & 0.24\cinf{0.02} & 3.40\cinf{0.25} \\
    \frshort \nogrammar & 0.41\cinf{0.06} & 1.49\cinf{0.18} \\
    \bottomrule
    \end{tabular}
    \caption{Means and standard deviations of the average absolute deviation from the gold length by model on ATIS, \ie lower is better}
    \label{tab:length-atis}
\end{table}

\begin{table}[t]
	\resizebox{\linewidth}{!}{
	\begin{tabular}{lrr}
		\toprule
		\textbf{Model} & \textbf{iid} & \textbf{Length} \\
		\midrule
		LSTM seq2seq \nogrammar & 81.53\cinf{0.77} & 43.86\cinf{2.93} \\
		Rel. Transformer \nogrammar & 81.53\cinf{0.44} & 34.84\cinf{5.28} \\
		BART-base \nogrammar & 90.54\cinf{0.45} & 65.29\cinf{1.01} \\
		\frshort \nogrammar & 73.80\cinf{1.62} & 54.42\cinf{1.25} \\
		\bottomrule
	\end{tabular}}
\caption{Mean and standard deviations accuracy on the development sets of the \textbf{ATIS} splits (without grammar-based decoding).}
\label{tab:atis-dev}
\end{table}
\begin{table}[t]
	\resizebox{\linewidth}{!}{
	\begin{tabular}{lrrr}
		\toprule
		\textbf{Model} & \textbf{Calendar} & \textbf{Document} & \textbf{Email} \\
		\midrule
		BART-base \nogrammar & 94.8\cinf{0.3} & 56.2\cinf{10.9} & 91.4\cinf{0.4} \\
		\frshort\nogrammar & 86.4\cinf{3.4} & 66.1\cinf{4.8} & 85.0\cinf{2.8} \\
		\bottomrule
	\end{tabular}}
\caption{Mean and standard deviations accuracy on the development sets of the \textbf{Okapi} splits (without grammar-based decoding).}
\label{tab:okapi-dev}
\end{table}

\end{document}